\pdfoutput=1

\documentclass[11pt]{article}

\usepackage{emnlp2021}

\usepackage{times}
\usepackage{latexsym, amssymb, mathtools}
\usepackage{comment}
\usepackage{booktabs}
\usepackage{enumitem}
\usepackage{placeins}
\usepackage{flushend}
\usepackage{multirow, graphicx}
\usepackage[ruled,noend,vlined]{algorithm2e}
\usepackage[disable]{todonotes} 
\usepackage[T1]{fontenc}

\usepackage[utf8]{inputenc}

\usepackage{microtype}

\title{Bandits Don't Follow Rules: \\Balancing Multi-Facet Machine Translation with Multi-Armed Bandits}

\author{Julia Kreutzer \and David Vilar \and Artem Sokolov\\
        Google Research\\
        \texttt{\{jkreutzer, vilar, artemsok\}@google.com}}

\begin{document}
\maketitle
\begin{abstract}
Training data for machine translation (MT) is often sourced from a multitude of large corpora that are multi-faceted in nature, e.g. containing contents from multiple domains or different levels of quality or complexity. Naturally, these facets do not occur with equal frequency, nor are they equally important for the test scenario at hand. 
In this work, we propose to optimize this balance jointly with MT model parameters to relieve system developers from manual schedule design. 
A multi-armed bandit is trained to dynamically choose between facets in a way that is most beneficial for the MT system.  
We evaluate it on three different multi-facet applications: balancing translationese and natural training data, or data from multiple domains or multiple language pairs. We find that bandit learning leads to competitive MT systems across tasks, and our analysis provides insights into its learned strategies and the underlying data sets.
\end{abstract}

\section{Introduction}
Parallel training data for machine translation (MT) is commonly sourced and combined from multiple large sub-corpora to obtain the maximum number of training examples. The WMT shared tasks, for example, provide a number of distinct training corpora since~\cite{koehn2006statistical}. Such corpora are \emph{multi-faceted} in nature, consisting of a generally unbalanced \emph{mixture of data sources that differ from each other} in word distribution, domain or other traits. Examples of such differences could range from strongly heterogeneous data like distinct languages for training multi-lingual systems~\citep{dong2015multi, firat-etal-2016,arivazhagan2019massively} to rather subtle variations in data provenance (e.g. human-generated vs.\ machine-produced data crawled from web), through a mid-strength variation in multi-domain MT~\cite{farajian2017multi,mueller2020domain,pham2021revisiting}. 
The nature of data facets and their identity is known at training time, either from the data sources directly, for example meta-data from data collection pipelines, or can be provided by dedicated classifiers---
but this important information is discarded when mixing and shuffling them for training~\citep{arjovsky2019invariant, teney2020unshuffling}. 
Thus, the optimal balance of facets needs to be decided beforehand, and with the requirements at test time in mind. At test time, facets may be equally important, but they might not all have the same amounts of training data. 
These data balancing decisions are time-consuming and expensive as they often require multiple iterations for striking the right balance between, on the one hand, robust performance at test time on underrepresented facets and, on the other hand, preserving valuable linguistic and lexical information contained in the higher-represented ones. For additional complication, potential positive and negative transfer between facets should be taken into account~\citep{arivazhagan2019massively,wang2021gradient}. The complexity of these decisions exacerbate as training data grows.

\begin{figure}[t]
    \centering
    \includegraphics[width=0.55\columnwidth]{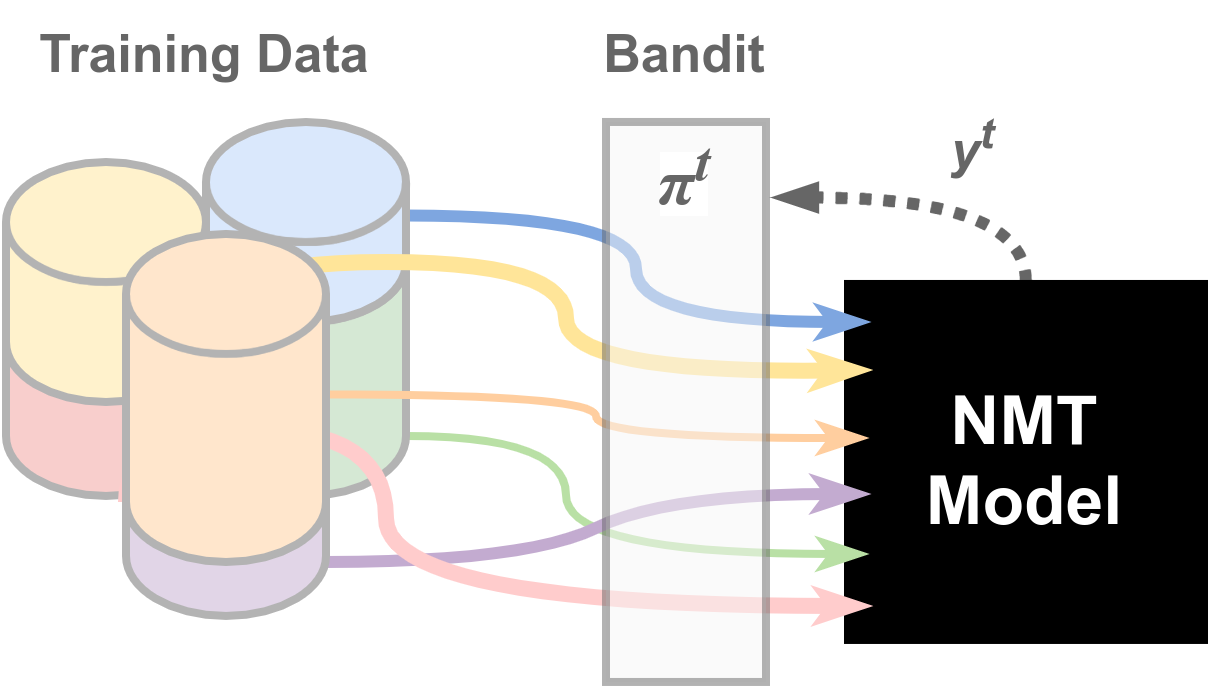}
    \caption{Multi-armed bandits for NMT data selection.}
    \label{fig:diagram}
\end{figure}

Even with established data balancing heuristics in place (e.g. upsampling with a tuned temperature $\tau$\footnote{Sampling from an annealed and renormalized empirical distribution over facets $f$, $p(f)=\text{\rm softmax}_{f}(\ln(\hat p(f))/\tau)$.}~\citep{devlin_multi_bert,arivazhagan2019massively}), 
different balances might be needed at different stages of training. This realization kick-started a development of training curricula~\citep{bengio2009curriculum} which, despite efforts in neural MT, have yet to produce a recipe applicable to concrete data at hand~\citep{zhang2018empirical}. 
Existing curricula presuppose fixed 
notions of difficulty and come with hand-crafted schedules, often inspired by the human learning process~\citep{kocmi2017curriculum, zhang2018empirical, platanios2019competence}. 
Such approaches are brittle in that they may not generalize well across tasks, and there has been evidence that even the reverse of the initially hypothesized order works well~\citep{bengio2009curriculum, wang2018denoising, zhang2018empirical}. This suggests that our human intuitions about difficulty and data succession may not correspond to the optimization process of an NMT system~\citep{demystify}.

In this paper we argue for \emph{automatically learned and adaptive data curricula}, where the learning system explicitly chooses a facet at each point in training, and does not depend on presupposed schedules. 
This has three major advantages: First, it relieves system developers from lots of manual work. Second, it can improve quality by ignoring irrelevant, redundant or already learned data. Third, it can directly optimize for a uniform performance objective to maintain quality on all facets. As a side effect, post-training analyses may improve data interpretability and efficiency~\cite{gasco2012does}.
However, outsourcing data decisions to an auxiliary ML model sacrifices some of control and understanding. In particular, multiple training data selection strategies can lead to models of comparable quality, especially when measured by crude metrics like BLEU.

We formulate \emph{multi-faceted training as a multi-armed bandit learning problem}, where the arms/actions correspond to the available facets in the training data. At each training step, the bandit chooses one facet for the MT system to train on and receives a reward signal whether this choice was beneficial for the training progress (see Figure~\ref{fig:diagram}). We implement the EXP3 algorithm~\cite{auer2002finite} as proposed for automated curriculum learning~\cite{graves2017automated} (\S\ref{sec:approach}), and evaluate it on three different multi-facet applications for machine translation. These require balancing training data that is natural or translationese (\S\ref{sec:nat2trans}), comes from a variety of domains (\S\ref{sec:domain}), or from many different languages (\S\ref{sec:multilingual}). To the best of our knowledge this is the first study that addresses these problems jointly and provides a competitive solution to all of them. We analyze the effects of different reward signals and chosen facets over time, shedding new light on the importance of different facets for each of the tasks. 



\section{Learning to Select Data with Bandits}\label{sec:approach}
Learning a data curriculum can be framed as a multi-arm bandit problem, where the decision to train on a particular subset of data is outsourced to a bandit algorithm that is learned alongside the main task~\citep{graves2017automated}. 
After the bandit chooses a facet, the NMT system is updated on a uniformly sampled batch of data from this facet. The system then provides a reward to the bandit, telling how successful this selected batch of data was in terms of overall training progress (see Figure~\ref{fig:diagram}). 


Formally, the bandit selects actions from a set $\mathcal{A}$ which is a discrete set of ids. In each round $t$, the bandit selects an action $a^t \in \mathcal{A}$ and observes a scalar loss, $y^t = Y^t_{a^t}$, where $Y^t$ is the complete but unobserved loss vector for each possible action. The bandit parameters are updated to minimize the regret $R = \mathbb{E}[\sum_t y^t] - \min_a \sum_t Y^t_a$ of not playing the arm that is best in hindsight.
We operate in a fully adversarial setup assuming that reward vectors $Y^t$ can be arbitrary, i.e.,~they can depend on the full history, data etc., but cannot be adaptive to the selected action~$a^t$. 

With a collection of subsets of training data (facets), 
covering the full training data, $\cup \mathcal{D}_a = D$, the EXP3 algorithm proceeds as follows \cite{auer2002finite, graves2017automated}:

\begin{algorithm}[h!]
\SetAlgoLined
\SetKwInOut{Input}{Input}
\Input{NMT model $\theta^0$, number of facets $n$, exploration rate $\gamma$, bandit learning rate $\mu$, training facets $\mathcal{D}_a$}
\DontPrintSemicolon
\LinesNumbered
\KwResult{Sequence of arms $\{a^1, a^2, \dots, a^T\}$}
 Initialize weights $\mathbf{w} = \mathbf{0} \in \mathbb{R}^n$\\
 \For{$t = 0, \dots, T$}{
  $\pi^t(a) \coloneqq (1-\gamma) \frac{\exp(\mathbf{w}_a)}{\sum_a \exp(\mathbf{w}_a)} + \frac{\gamma}{n}$\;
  sample $a^t \sim \pi^t$\;
  sample a batch $B^t$ uniformly from $\mathcal{D}_{a^t}$\;
  NMT update step on $B^t$ to get $\theta^{t+1}$\;
  measure learning progress $y^t$\;
  update $\mathbf{w}_{a} = \mathbf{w}_a + \mu y^t[\![a = a^t]\!]/\pi^t(a)$\;
 }
 \caption{Multi-Facet EXP3 for NMT}
 \label{algo}
\end{algorithm}

The regret $R$ behaves as $O(\sqrt{T \ln d})$ \cite{auer2002finite}, so in the limit the bandit will do as good as the best arm from $\mathcal{A}$, i.e.,\ $R/T\rightarrow 0$ as $T\rightarrow\infty$. 
\citet{graves2017automated} 
used a slightly modified algorithm EXP3.S~\cite{auer2002finite} that competes against any \emph{sequence} of actions to reflect dynamic changes. In practice, we found the performance of the vanilla EXP3 sufficient for NMT.

\paragraph{Measuring learning progress} \citet{graves2017automated} propose a variety of reward functions for measuring learning progress. In this work, we focus on rewards that are functions of the loss value: 
\begin{itemize}[leftmargin=0.2in]\itemsep0em 
    \item \texttt{loss}: the plain loss objective value, $\mathcal{L}(\theta^t)$; 
    \item \texttt{pg}: absolute prediction gain, $\mathcal{L}(\theta^t)-\mathcal{L}(\theta^{t+1})$;
    \item \texttt{pgnorm}: relative \texttt{pg}, $1-\mathcal{L}(\theta^{t+1})/\mathcal{L}(\theta^t)$.
\end{itemize}

They can be evaluated
on the training batch $B^t$ or on a development batch $B^t_{dev}$ (denoted with prefix \texttt{dev-}). Due to the ambiguity of what can lead to high losses on the training set---such as noisy, unseen or untypical examples---rewards calculated on batches sampled from dev sets (that in our work contain an equal mix of facets) proved to be more successful in our experiments. This also allowed us to inject the equal importance of facets into the rewards. We calculate rewards on randomly sampled batches, since it is cheaper than on the full dev set~\citep{kumar2019reinforcement}. This is not an issue for EXP3, as it allows rewards to be non-deterministic: it provably converges for adversarial feedback, and so for identically distributed random rewards too. We linearly re-scale the rewards to $[-1;1]$ clipping them between the 20th and the 80th quantiles of the most recent 5k rewards~\citep{graves2017automated}. 

\paragraph{Training costs} The memory overhead of training the bandit alongside the main NMT is negligible since it only requires the storage of reward and sampling statistics across arms (see Algorithm~\ref{algo}). There is no overhead in terms of speed for the \texttt{loss} reward since it is already computed during the normal MT training, but a second forward pass is needed to calculate prediction gains (\texttt{pg} or \texttt{pgnorm}). With the cost of a forward pass $c$, the additional computational cost per iteration is $O(c)$ for one evaluation batch, independent of the number of facets. This is notable cheaper than recent methods based on gradient similarity that require backward passes on training and development sets for each facet, plus a gradient update for a parametrized policy~\cite{wang2020optimizing}.

\section{Experiments}

\paragraph{Data}
To ensure that our recipe generalizes to multiple setups
we empirically tested our approach on three different tasks varying across several dimensions (Table~\ref{tab:data}): 
\begin{enumerate}[leftmargin=0.2in]\itemsep0em 
\item \textbf{Natural vs.~translationese}: For large-scale \texttt{en-de} translations we model two facets with a subtle distinction, namely the distinction of ``translationese'' and ``natural'' target sides (\S\ref{sec:nat2trans}). The difficulty lies in the weak demarcation between classes of signals, large provenance diversity of data and the overall large data size.
\item \textbf{Multi-domain}: We train a multi-domain NMT system for \texttt{en-de} with mid-size training data (\S\ref{sec:domain}). The automated curriculum has to balance facets of the same language, but with subtle domain-specific differences.
\item \textbf{Multilingual}: We experiment with multilingual NMT models on two small-scale subsets of 8 language pairs from the TED57 dataset and the large-scale OPUS100 set with 99 language pairs (\S\ref{sec:multilingual}). 
Facets are defined as language pairs and they are related to varying degrees, so reward signals are expected to vary in terms of dynamics and strength.
\end{enumerate}
\begin{table}
    \centering
    \resizebox{\columnwidth}{!}{
    \begin{tabular}{l|cccc}
    \toprule
        \textbf{Corpus} & \textbf{Lang. pairs} & \textbf{Facets} & \textbf{Entropy} &\textbf{Sent.} \\
        \midrule
        Nat.-Transl. & 1 & 2 & 96.9\% & 57M \\ 
        Multi-domain & 1 & 5 & 85.4\% & 1.5M \\ 
        TED57 diverse& 8 & 8 & 78.9\% & 766k \\ 
        TED57 related& 8 & 8 & 73.1\% & 586k \\ 
        OPUS100 M2O/O2M & 99 & 99 & 91.6\% & 55M\\  
        OPUS100 M2M & 198 & 198 & 92.7\% & 109M \\
         \bottomrule
    \end{tabular}
    }
    \caption{Overview of multi-faceted training data sets. The entropy of the frequency distribution of facets as present in the corpus is measured in percents of the maximum natural entropy. } 
    \label{tab:data}
\end{table}

\paragraph{Implementation}
\begin{table*}[]
    \centering
    \resizebox{\textwidth}{!}{
    \begin{tabular}{ll|c|c|cc|c|cc}
        \toprule
        & & \multirow{2}{*}{Avg} & Translationese & \multicolumn{2}{c|}{Natural} & Translationese & \multicolumn{2}{c}{Natural}\\
        & & &\textbf{WMT20} & \textbf{WMT20-paraph} & \textbf{WMT20-rev} & \textbf{WMT18} & \textbf{WMT18-paraph} & \textbf{WMT18-rev} \\
        \midrule
        \midrule
        & Baseline & 26.42 &27.64 & 9.23 &22.87 & 52.19 & 12.61& 34.00 \\
        & Tagged & 27.12 & 28.05 (29.37) & 9.96 & 23.92 & 52.24 (50.85) & 13.12 & 35.44 \\
        \midrule
        \parbox[t]{2mm}{\multirow{6}{*}{\rotatebox{90}{Bandit}}} & \texttt{loss} & 27.37 & 27.81 & 9.37&24.36&50.34& 12.73 & 39.61  \\
        & \texttt{pg} & 27.66 & 28.32 & 9.48 & 24.66 & 51.52& 12.96 & 39.03  \\
        & \texttt{pgnorm} & 26.40& 27.49 & 9.30 & 22.42 & 51.37  & 12.33 & 35.48\\
        & \texttt{dev-loss} &27.47 &28.19 &9.49 &23.99 & 51.57& 12.64 & 38.94  \\
        & \texttt{dev-pg} & 27.39&27.53 &9.28  & 24.56 &51.18   &12.67& 39.12  \\
        & \texttt{dev-pgnorm} & 27.22&27.68 & 9.40 &24.35&50.21& 12.67 & 38.98    \\
        \midrule
       \parbox[t]{2mm}{\multirow{3}{*}{\rotatebox{90}{CDS}}} & Baseline+CDS & 27.74&29.52 & 9.76 & 23.85 &53.62& 12.99 & 36.71 \\
        & Tagged+CDS & 27.50 & 29.17 (28.98) & 10.00 & 23.58 & 53.60 (50.18) &{13.17} & 35.46 \\
        & \texttt{dev-pgnorm}+CDS & {28.01} &29.09 & 9.72 &{24.20} & 53.60 & 13.07 & {38.44} \\
        \bottomrule
    \end{tabular}
    }
    \caption{WMT \texttt{en-de}: BLEU scores on the natural vs.~translationese task. Source tags for tagged baselines correspond to the test set's  facet; for translationese sets, BLEU for the natural tag is in brackets.
    }
    \label{tab:wmt_ende}
\end{table*}
We implemented the Transformer model~\cite{vaswani2017all} in JAX~\cite{jax2018github}, using the neural network library Flax~\cite{flax2020github} (more details in \S\ref{sec:hapax}).
After training we select the model for testing that obtained the highest SacreBLEU score~\citep{post2018call} on development sets containing a balanced selection of all facets. 

\section{Results}\label{sec:results}
For each task we evaluate whether the bandit-directed training schedules can outperform the zero-effort ``take-it-all'' approach where datasets are concatenated and training examples are presented in random order. In addition, we compare it to task-specific best practices, and investigate which strategies are learned by the bandit schedules.

\subsection{Natural vs.~translationese NMT}\label{sec:nat2trans}

\paragraph{Setup} 

We train a \texttt{big} Transformer on the concatenation of the News Commentary (v15), ParaCrawl (v5.1), Europarl (v10) and CommonCrawl training corpora.\footnote{\href{www.statmt.org/wmt20/translation-task.html}{WMT2020 news translation task}.}
Since natural vs.~translationese facets are not explicitly marked in the corpora, 
we train two neural LMs for the target language, one on natural text and one on translated text, and select the higher-scoring one as label (\S\ref{sec:hyperparams}). We are interested in improving the naturalness of the translation output, but this is hardly measured by automatic metrics, because standard reference translations are translationese, so BLEU might even give contradictory signals~\citep{freitag2020bleu}.
Therefore, 
we also evaluate on the \emph{reverse} direction WMT20 set, i.e.
the reversed test set for \texttt{de-en} (suffix `-rev'), which consist of original German text.
This serves as a proxy for measuring the naturalness of the system output.
We additionally use the references provided by~\citet{freitag2020bleu} which were paraphrased versions of the official ones, with the goal of improving their naturalness (denoted with the suffix `-paraph' in the results).
The bandit development set contains 2000 sentences
of equal mixture of natural (the `rev' part) and translationese sentences from the WMT19 newstest, and the final evaluation is on faceted WMT18 and WMT20 news test sets.

As a simple controlled translation, we also train tag-based baselines~\citep{riley-etal-2020}, where source tags correspond to facets, also during testing. As the natural mode is what often desired, we additionally evaluate translations with the `natural' tag for translationese sets.

\paragraph{Results} All bandit approaches improve over the baseline (Table~\ref{tab:wmt_ende}) by around 0.5--0.9 BLEU on average across test sets (except for \texttt{pgnorm}), but the individual tendencies vary across reward choices. 
The dynamics of the bandit arm probabilities~(\S\ref{app:probs}, Figure~\ref{fig:wmtbase_ende_base_probs}) reveal that most rewards prefer the natural part of data since it is harder to learn for the NMT system and results in consistently higher loss values; except for \texttt{pgnorm}, which also loses on the reverse set. 
Additional data filtering by Contrastive Data Selection (CDS)~\citep{wang2018denoising} leads to major improvements for the baseline on translationese and natural test sets.
This approach filters the training data by removing 30\% of sentences that are considered noisy by a model iteratively trained on trusted data (here NewsCommentary v15).
It was trained independently of the natural and translationese distinction, so the CDS improvements are due to a generally improved quality of the training data.
It strengthens bandit results in a similar way,
gaining about 0.3 and 1.7 BLEU on two natural tests set while performing comparably on the others, which shows that both approaches are complimentary---we speculate that CDS removing noisy examples allows bandits to better focus on truly difficult examples.
Comparing the \texttt{dev-pgnorm} bandit without CDS and the baseline with CDS on natural `rev' test sets suggests that the bandit could compensate the lack of data filtering.

\begin{table*}[t]
    \centering
    \resizebox{0.8\textwidth}{!}{
    \begin{tabular}{ll|c|ccccc}
        \toprule
        & & Avg & \textbf{Med} & \textbf{IT} & \textbf{Law} & \textbf{Koran} & \textbf{Subs} \\
        & Size (k) & 291.2 & 248.0 & 467.3 & 222.9 & 18.0 & 500.0\\
        \midrule
        \midrule
        \multirow{2}{*}{\rotatebox{90}{Base}} & \citet{aharoni2020unsupervised} &40.2& 53.3 & 42.1 & 57.2 & 20.9 & 27.6 \\
                & Ours &39.42& 51.65 & 41.47 & 55.08 & 21.24 & 27.67 \\
        \hline
        \multirow{4}{*}{\rotatebox{90}{Static}} &  Uniform ($\tau=\infty$) &39.26& 51.91 & 44.24 & 50.93 & 22.34 & 26.90\\
       & Upsampled ($\tau=5$) & 38.78 & 51.11 & 41.10 & 52.87 & 23.07 & 25.72\\
        & Proportional ($\tau=1$) &40.40&52.83 &44.65 & 54.53& 21.23&28.76\\
        & Inverse Proportional ($\tau=-1$) &40.03& 53.43 & 44.97 & 51.84 & 22.70 & 27.20\\
        \hline
        \parbox[t]{2mm}{\multirow{6}{*}{\rotatebox{90}{Bandit}}} & \texttt{pg} &40.26& 53.38 & 46.42 & 51.44 & 22.19 & 27.85\\
        & \texttt{pgnorm} &38.78& 51.76 & 45.97 & 50.03 & 21.43 & 24.72\\
        & \texttt{loss} &39.12& 50.63 & 42.89 & 51.19 & 23.08 & 27.83\\
        & \texttt{dev-loss} &39.96& 50.22 & 42.48 & 55.68 &23.34  & 28.06\\
        & \textbf{\texttt{dev-pg}} &40.56& 53.35 & 42.66 & 55.95& 22.62 & 28.23\\
        & \textbf{\texttt{dev-pgnorm}} &40.56& 53.23 & 42.99 & 55.89 & 22.79 & 27.92\\
        \bottomrule
    \end{tabular}
    }
    \caption{Test BLEU scores on the multi-domain task. Rewards in bold improve over the baseline uniformly.} 
    \label{tab:multidomain_detok}
\end{table*}

\subsection{Multi-Domain NMT}\label{sec:domain}

\paragraph{Setup} We follow the multi-domain setup for German-to-English translation by \citet{mueller2020domain} using the data re-split by \citet{aharoni2020unsupervised}. By construction it contains in-domain data from five domains and no auxiliary general-domain data, thus preventing data augmentation with pseudo in-domain data selection~\cite{axelrod2011domain}.  The goal of this evaluation is to improve uniformly on all domains using a mixed training set. As in prior work, we use the \texttt{base} Transformer architecture, and, where possible, try to match the training setup from \citep{aharoni2020unsupervised} (\S\ref{sec:hyperparams}). However, we were not able to exactly replicate their scores due to inevitable implementation differences. 

\begin{figure}
    \centering
    \includegraphics[width=0.9\columnwidth]{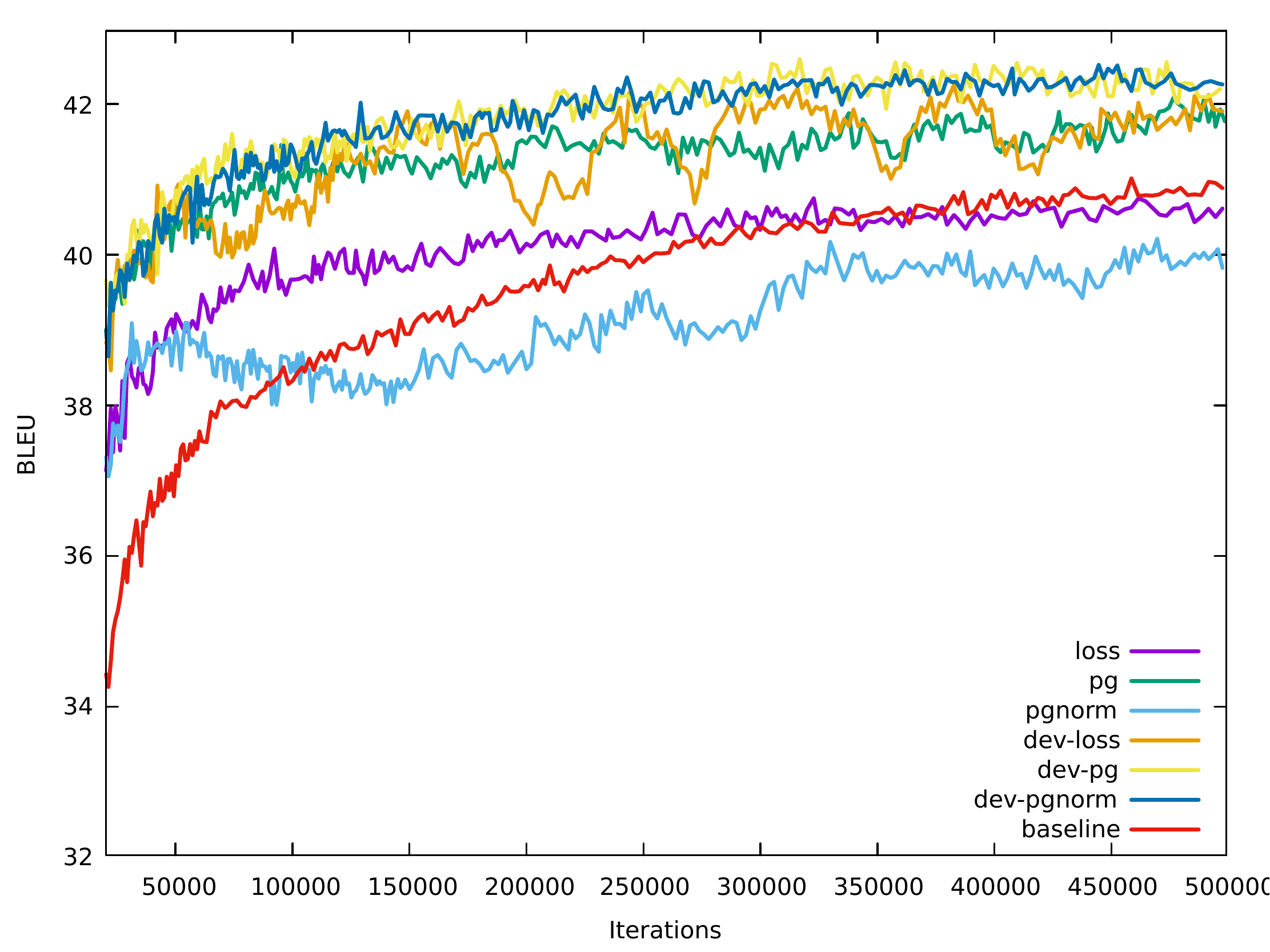}
    \caption{Evaluation scores on the mixed development set during training for the multi-domain task.}
    \label{fig:multi_domain_curves}

\end{figure}

\paragraph{Results} The two most successful bandit data selection strategies (\texttt{dev-pg} and \texttt{dev-pgnorm}) converge faster than the baseline (Figure~\ref{fig:multi_domain_curves}) \emph{and} achieve better scores (up to +1.7 point above the baseline on some domains and 1.1 points on average). Analysing the evolution of facet sampling probabilities 
(\S\ref{app:probs}, Figure~\ref{fig:multi_domain_probs}), we
find that \texttt{dev-pg} and \texttt{dev-pgnorm} focused largely on Law and Subtitles domains. We hypothesize that these rewards are capitalizing on the higher sentence quantity and hence potential diversity of the higher-resource domains. At the same time, they quickly neglect the IT and Koran domains, which may be structurally simple and/or monotonic. Not frequently training on examples from latter domains does not lead to a decrease of translation quality on them. In general, gains in quality over the baseline are not related to the sampling preferences of the bandits. This highlights the difficulty of designing a proper schedule manually and prior to training using intuition only. Static temperature-based sampling yields gains tied to the availability of resources, 
(e.g.~improvements for $\tau=1$ on the high-resource domains, and $\tau=-1$ on the low-resource domains, except for $\tau =5$ which gains only for Koran), but they---in contrast to the dynamic bandits---fail to improve on all domains. This shows that the additional flexibility of the bandits to adapt the sampling distribution during training is beneficial for equitable quality gains.

\subsection{Multilingual NMT}\label{sec:multilingual}

\paragraph{Setup}
In multilingual MT, parallel training data is often paired with English, so there are three major training setups for multilingual translation: many-to-one (M2O), learning to translate many languages into English; one-to-many (O2M), translating from English, and many-to-many (M2M).
We experiment with M2O translations for the \texttt{diverse} and \texttt{related} subsets of the multilingual TED dataset~\citep{qi2018when}.
The two subsets cover 8 languages with very different data sizes, selected as pairs of related languages of different size~\citep{neubig2018rapid} or a set of diverse languages from different language families and with different scripts~\citep{wang2020balancing}. There are large discrepancies in the sizes of the subsets for each language, e.g. \texttt{be} has only 4.5k sentences, while the related \texttt{ru} has 208.4k. This makes it valuable for testing the behavior of the bandit with facets that are linguistically similar but very differently scaled. 
For a more data-balanced setup, we experiment with the OPUS100 dataset~\citep{zhang2020improving}, which contains up to 1M of training examples sampled from the entirety of the OPUS collection of parallel corpora from various domains~\citep{opus} for 99 languages paired with English, of which 94 come with test sets. As a result, the data has large inter- and intra-facet diversity.
For both evaluation scenarios we train SentencePiece models~\cite{kudo2018sentencepiece} on a re-balanced corpus~\citep{nguyen2017transfer,m2m}\footnote{Upsampling all languages to the maximum size.} to create a vocabulary of 32k tokens, add target language tags and train Transformer \texttt{base} models. We construct a balanced development set by randomly selecting a fixed number of sentences from the language-specific development sets (500 for TED; 100 for OPUS) to reflect our interest in high quality across all languages. Rewards for the bandit are computed on samples from this balanced dataset. 
We compare with static uniform sampling distributions ($\tau=\infty$) over facets, and size-proportional ($\tau=1$) or upsampled ($\tau=5$) distributions, since they have been reported successful in previous works~\citep{wang2020balancing, zhang2020improving}. They sample batches of a single language at each step, while the vanilla baseline samples mixed-language batches from the shuffled concatenated data.\footnote{The literature has been divided whether to mix batches \citep{aharoni2019massively, zhang2020improving, zhang2021share, demystify} or not~\citep{firat-etal-2016, wang2020balancing}.} All other hyperparameters can be found in \S\ref{sec:hyperparams}.
We report experiments with the \texttt{dev-pgnorm} reward since it performed best.

\begin{figure}[t]
    \centering
    \includegraphics[width=0.8\columnwidth]{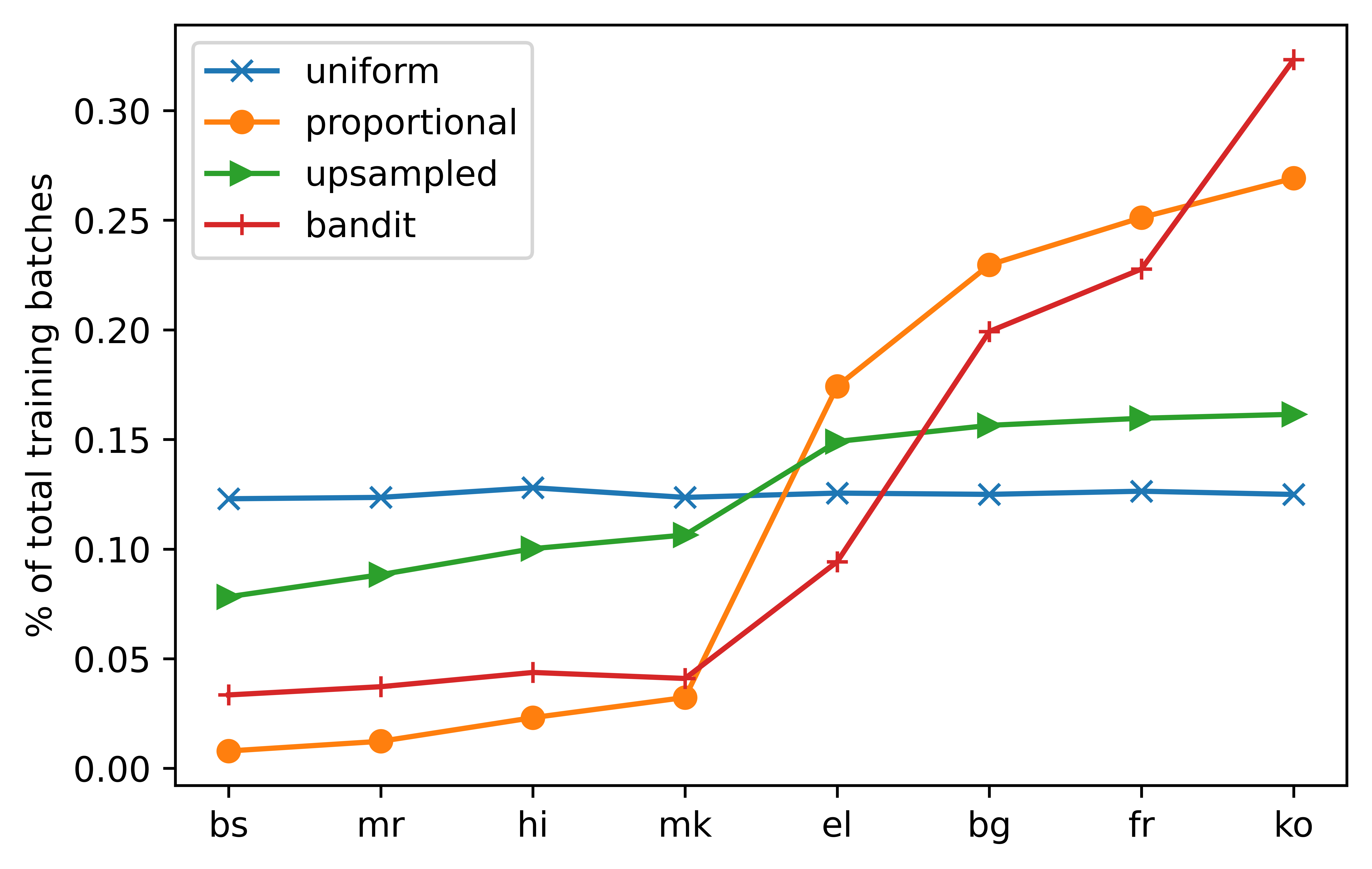}
    \caption{Total number of batches trained on for each language throughout training for TED-diverse M2O.}
    \label{fig:probs_m2o_diverse}

\end{figure}

\paragraph{TED Results}
Tables~\ref{tab:diverse_m2o} and~\ref{tab:related_m2o} compare our results on the \texttt{diverse} and \texttt{related} subset with the most recent work of~\citet{wang2020balancing}, who proposed a dynamic data scheduling algorithm (MultiDDS) based on gradient similarity between training and development data.
On average, our implementations of batch-wise uniform or proportional sampling yield similar results to theirs for the \texttt{diverse} set, but on the \texttt{related} set they perform slightly worse, because~\citet{wang2020balancing} train on more data for \texttt{sl} (61.5k) and \texttt{pt} (185k) than is contained in the publicly available dataset, resulting in a difference of 8 and 5 BLEU on respective languages.\footnote{Downloaded from \url{https://github.com/neulab/word-embeddings-for-nmt}.}
The bandit consistently outperforms the mixed-batch baseline (`Base') and performs similarly to proportional sampling of language-specific batches (`Proportional'). Trivially sampling languages according to their size is a good heuristic for both setups despite the large data discrepancies between languages, corroborating previous findings on this dataset~\citep{neubig2018rapid,demystify}. It yields better results than a uniform sampling scheme (and than the commonly used $\tau=5$) which a practitioner might have chosen without prior knowledge about the task.
The bandit automatically discovers this insight without having access to explicit size information, as can be seen in
Figure~\ref{fig:probs_m2o_diverse} for the diverse set (related: \S \ref{app:strategies}, Figure~\ref{fig:ted_related_plays}). 
Compared to size-proportional sampling, it slightly upsamples all smaller languages 
and slightly downsamples some of the larger ones (\texttt{el}, \texttt{bg}, \texttt{fr}), but not as strongly and consistently as the $\tau=5$ upsampling.
This led to an improvement of the translation quality of the lowest-resource languages, and was incentivized by the equal presence of languages in the balanced development set used for reward calculations. With a similar incentive but much more expensive updates, Multi-DDS's gains over proportional sampling are also on the smallest datasets.

\begin{table*}[t]
    \centering
    \resizebox{0.9\textwidth}{!}{
    \begin{tabular}{ll|c|cccccccc}
    \toprule
     &    & Avg & \texttt{\textbf{bs}} & \texttt{\textbf{mr}} & \texttt{\textbf{hi}} & \texttt{\textbf{mk}} & \texttt{\textbf{el}} & \texttt{\textbf{bg}} & \texttt{\textbf{fr}} & \texttt{\textbf{ko}} \\
      &  Size (k) & 95.8 & 5.7 & 9.8 & 18.8 & 25.3 & 134.3 & 174.4 & 192.3 & 205.6 \\    
    \midrule
    \midrule
           Ours  & Base& 25.43 & 21.59 & 10.00 & 20.73 & 29.90 & 33.74 & 34.73 & 35.70 & 17.06 \\ 
    \midrule
        \multirow{4}{*}{\citet{wang2020balancing}}  & Uniform ($\tau=\infty$) & 24.81 & 21.52 & 9.48 & 19.99 &  30.46 & 33.22 & 33.70 & 35.15 & 15.03 \\
      & Upsampled ($\tau=5$) & 26.01 & 23.47 & 10.19 & 21.26 & 31.13 & 34.69 & 34.94 & 36.44 & 16.00 \\
      & Proportional ($\tau=1$) & 26.68 & 23.43 & 10.10 & 22.01 & 31.06 & 35.62 & 36.41 & 37.91 & 16.91 \\ 
      &  MultiDDS-S & 27.00 & 25.34 & 10.57 & 22.93 & 32.05 & 35.27 & 35.77 & 37.30 & 16.81 \\ 
   \midrule 

       \multirow{3}{*}{Static} & Uniform ($\tau=\infty$) & 24.47 & 21.72 & 8.13 & 17.84 & 29.18 & 33.37 & 34.22 & 35.99 & 15.31 \\
       & Upsampled ($\tau=5$) & 26.04 & 24.60 & 9.56 & 19.68 & 30.81 & 34.71 & 35.54 & 36.74 & 16.68 \\
        & Proportional ($\tau=1$) & \textbf{26.85} & 22.00 & 9.73 & \textbf{21.02} & \textbf{32.57} & \textbf{36.27} & \textbf{37.37} & \textbf{38.29} & \textbf{17.57} \\
    \midrule

      Bandit & \texttt{dev-pgnorm} &  26.30 & \textbf{23.88} & \textbf{10.41} & 20.70 & 31.18 & 34.10 & 35.87 & 36.81 & 17.47 \\
    \bottomrule
    \end{tabular}
    }
    \caption{BLEU on \texttt{diverse} TED data for many-to-one models.}
    \label{tab:diverse_m2o}
\end{table*}

\begin{table*}[]
    \centering
    \resizebox{0.9\textwidth}{!}{
    \begin{tabular}{ll|c|cccccccc}
    \toprule
        & & Avg & \texttt{\textbf{az}} & \texttt{\textbf{be}} & \texttt{\textbf{gl}} & \texttt{\textbf{sl}} & \texttt{\textbf{tr}} & \texttt{\textbf{ru}} & \texttt{\textbf{pt}} & \texttt{\textbf{cs}} \\
       &  Size (k) & 73.2 & 5.9 & 4.5 & 10.0 & 19.8* &  182.5 & 208.4 & 51.8* & 103.1 \\
    \midrule
    \midrule
        Ours & Base & 22.87 & 11.91 & 17.19 & 26.64 & 22.56 & 22.66 & 21.83 & 34.87 & 25.17 \\
        \midrule
         \multirow{4}{*}{\citet{wang2020balancing}} & Uniform* ($\tau=\infty$) & 22.63* & 8.81 &14.80 &25.22 & 27.32*& 20.16 &20.95 &38.69* &25.11\\
         & Upsampled* ($\tau=5$) & 24.00* & 10.42 & 15.85 & 27.63 & 28.38* & 21.53 & 21.82 & 40.18* & 26.26 \\
         & Proportional* ($\tau=1$) & 24.88* &11.20& 17.17 &27.51& 28.85* &23.09 &22.89 &41.60* &26.80\\ 
         & MultiDDS-S* & 25.52* &12.20 &19.11 &29.37 &29.35* &22.81 &22.78 &41.55* &27.03\\ 

    \midrule
       
        \multirow{3}{*}{Static} & Uniform ($\tau=\infty$) & 20.30 &  8.10 & 12.09 & 24.35 & 19.21 & 20.53 & 20.22 & 33.99 & 23.95\\
         & Upsampled ($\tau=5$) & 21.92 & 9.71 & 15.14 & 26.18 & 20.84 & 21.79 & 21.18 & 35.31 & 25.18 \\
        & Proportional ($\tau=1$) & \textbf{23.60} & 11.88 & 15.80 & 27.69 & \textbf{22.90} & \textbf{23.73} & \textbf{22.90} & \textbf{37.38} & \textbf{26.49} \\
    \midrule 

       \multirow{1}{*}{Bandit} & \texttt{dev-pgnorm} & 23.51 & \textbf{12.18} & \textbf{18.00} & \textbf{27.76} & 21.76 & 23.36 & 22.72 & 36.51 & 25.82 \\
    \bottomrule
    \end{tabular}
    }
    \caption{BLEU on \texttt{related} TED data for many-to-one models. *Trained on larger data than publicly available.}
    \label{tab:related_m2o}
    	
\end{table*}

\paragraph{OPUS100 Results} We compare against the M2O and O2M benchmark results set by
\citet{zhang2021share}, averaging results for languages with less than 0.1M sentence pairs (`Low'), those with 1M (`High'), and medium-sized ones (`Med'). \citet{zhang2021share} also use a Transformer \texttt{base}, but report averaged results for the last 5 checkpoints and create unbalanced vocabularies of twice the size of ours, resulting in a higher-capacity model. Our baselines therefore score slightly below, but the upsampling results are surprisingly competitive, which might be caused by implementation differences (e.g. upsampling the data before sampling mixed batching vs modifying the batch sampling distribution with homogeneous batches (ours)). 
The bandit clearly outperforms the vanilla baseline and static size-proportional sampling in both directions, and for M2O also uniform sampling, as reported in Table~\ref{tab:opus100}. It performs slightly weaker than the static upsampling approach. Uniform sampling is competitive for O2M, because it evenly balances the target language occurrence. M2M bandits improve over the baseline as well, on average +0.6 for M2O and +1.2 for O2M (\S\ref{app:m2m}), with the largest gain of +3.6 BLEU on O2M for the lowest-resource languages.

There is no correlation between training data size and BLEU on the test set for the baseline, nor between the sampling frequencies of the bandit and training data size for any of the directions (in contrast to the TED experiments). The bandits pursue selective strategies with very frequent switches between facets.  For M2O 11\% of all training steps were done on \texttt{nl}, and more than half the languages were sampled in less than 0.5\% steps each. For O2M, samples from \texttt{fy}, \texttt{ga}, \texttt{ky}, \texttt{mg} and \texttt{ug} were used in more than 3\% of steps each, and again around half the languages were trained on for less than 0.5\% steps. Comparing M2O and O2M top-5 sampled languages, we find 4 of those to be high-resourced (1M training examples) for M2O, but for O2M these are all mid to low-resourced with 27k-591k examples (details in \S\ref{app:strategies}). Surprisingly, the languages which are rarely sampled do not stand out with low translation quality. The selection of domains for the data sets is not controlled for in this benchmark~\citep{zhang2021share}, so we suspect domain effects might be interfering with BLEU reporting, in that some test sets might be more specialized than others, especially low-resource languages which are mainly covered by technical or religious data sets in OPUS. We suspect that this domain interference also makes it harder for the bandits to pick up reliable reward statistics per language, since there will be a lot of variance in terms of complexity within each language.

\begin{table*}[]
    \centering
    \resizebox{0.9\textwidth}{!}{
    \begin{tabular}{ll|cccc|cccc}
    \toprule
        & & \multicolumn{4}{c|}{\textbf{M2O}} & \multicolumn{4}{c}{\textbf{O2M}} \\
        & & All & Low & Med & High & All & Low & Med & High \\
    \midrule 
    \midrule
    
   Ours & Base & 28.41 & 29.53 & 27.54 & 28.44 & 19.78 & 19.72 & 18.65 & 20.51 \\
   \midrule
   \multirow{2}{*}{\citet{zhang2021share}} & Base & 29.27 & 29.71 & 30.10 & 28.55 & 20.93 & 18.02 & 22.36 & 21.39 \\ 
   & Upsampled ($\tau=5$) & 29.71 & 34.26 & 30.69 & 26.98 & 22.41 & 25.27 & 24.22 & 19.95 \\
    \midrule
       \multirow{3}{*}{Static} & Uniform ($\tau=\infty$) & 29.06 & 31.55 & 27.52 & 28.85 & 22.07 & 23.68 & 19.91 & 22.67 \\
      & Upsampled ($\tau=5$) & \textbf{30.15} & \textbf{32.72} &\textbf{ 28.48} & \textbf{30.00} & \textbf{22.35} & \textbf{23.81} & \textbf{20.46} & \textbf{22.86}\\
      & Proportional ($\tau=1$) & 28.07 & 29.70 & 27.29 & 27.80 & 19.39 & 18.84 & 18.05 & 20.49\\
      \midrule
       Bandit & \texttt{dev-pgnorm} & 29.53 & 31.64 & 27.73 & 29.66 & 20.30 & 21.77 & 18.23 & 20.91 \\ 
    \bottomrule
    \end{tabular}%
    }
    \caption{Average BLEU across languages pair groups for M2O \& O2M models evaluated on OPUS100 test sets.} 
    \label{tab:opus100}
\end{table*}

 \section{Related Work}\label{sec:related}

\paragraph{Model-based data selection} 
\citet{wees2017dynamic} reported first empirical success of hand-crafted schedules for data from different domains which are chosen according to cross-entropy scores of RNN-NMT models.
\citet{wang2018denoising} proposed an online data denoising approach, where 
noise is measured as the difference of log-probabilities between a learning model and the same model fine-tuned on small set of trusted data. Batches are composed of sentences with the highest contrastive data scores (CDS) corresponding to the least noisy sentences.
Our approach is similar to the above in that the multi-armed bandit acts on the online learning success of the MT model, but it is significantly cheaper since it does not require contrastive models nor a pre-defined schedule. Furthermore, the requirement of trusted data is lifted.

\paragraph{Difficulty-based curricula}
\citet{kocmi2017curriculum} apply the idea of curriculum learning~\citep{elman1993learning,bengio2009curriculum} to RNN NMT by simple ordering data in buckets corresponding to increased difficulty. 
\citet{zhang2018empirical} combine non-reusable buckets of difficulties with a manual schedule and achieve small improvements on small data with RNNs. \citet{platanios2019competence} apply a competence-based schedule with lengths and rarity to Transformer NMT that re-samples already used examples as long as they fall under the current competency. Many works on manually designed curricula note that presenting examples in the reverse order (hard-to-easy) works comparably well~\citep{bengio2009curriculum, wang2018denoising, zhang2018empirical}, which may be a sign of flawed intuitions. Our proposed solution groups data into facets rather than difficulty levels and reveals counterintuitive but effective schedules.

\paragraph{Learned curricula} 
Apart from \citep{graves2017automated}, whose curriculum learning bandits we adapt for NMT, \citep{kumar2019reinforcement} is closest to our work.
They frame the data selection task as an RL problem and define actions as data clusters corresponding to bins of CDS scores~\citep{wang2018denoising}. 
The same idea of representative batches is reused for multi-armed bandits enhanced with state representations in~\citep{kumar2021learning}. In \cite{wang2020optimizing} another RL algorithm is deployed for optimizing a distribution over training examples using the alignment of training and development gradients as rewards, requiring two backward passes on every step and an additional forward pass on an auxiliary neural net. In contrast to the RL-based approaches, we use light-weight bandits without state representations, which reduces memory and time complexities drastically.  

\paragraph{Bandit learning in MT} Multi-armed bandits were used in MT to improve general quality, either from online simulated user feedback~\citep{sokolov2015bandit,sokolov16stochastic, sokolov2017shared,kreutzer2017bandit,kreutzer2018reliability} or from offline logs~\citep{lawrence2017counterfactual,kreutzer-etal-2018} for domain adaptation. \citet{naradowsky2020machine} applied bandit algorithm to select the best NMT system for a particular translation task, when maintaining of multiple such systems is possible. More generally, RL approaches also seek to improve quality by focusing on more task-informed objectives~\citep{shen2016minimum} and improved approximations to the NMT policies~\citep{bahdanau2017actor}. Unlike these approaches, we treat the NMT model as a black box and do not intervene with its inner workings (see Figure~\ref{fig:diagram}).

\paragraph{Translationese vs.~natural MT}
\citet{toral2018attaining} have shown that the original language a sentence has been written in has a big impact on translation quality, i.e.,\ translating a sentence originally written in the source language is more difficult than translating (back) a sentence that was originally written in the target language and then translated into the source language. 
This second condition is `unnatural' for the actual use case of translation systems, but occurs frequently in translation evaluations, if the same dataset is used for evaluating both translation directions.
To avoid such artifacts, source sentences for evaluation should be been written originally in the source language~\cite{barrault2019findings}. Recently, \citet{vanmassenhove2021machine} showed that MT outputs present lower lexical diversity than human produced texts. MT systems generating outputs closer in style to the original target text are preferred by human judges~\citep{freitag2020human}. Hence our motivation (\S\ref{sec:nat2trans}) to produce more natural sounding translations.

\paragraph{Multi-Faceted MT} Multi-task learning~\citep{caruana1997multitask} for NMT was introduced by ~\citet{luongmtl} with the motivation to support a primary tasks with auxiliary data from related tasks. When understanding languages as tasks~\citep{dong2015multi, firat-etal-2016}, one single MT model can be used to translate between a multitude of languages and in particular also between translation pairs that were not in the training set~\citep{johnson-etal-2017,aharoni2019massively}. 
To address the problem of data imbalance~\citet{devlin_multi_bert,arivazhagan2019massively} proposed temperature sampling to upsample low-resource languages and downsample higher-resource ones (with $\tau>1$), that has since turned into the go-to data weighting strategy~\citep{freitag-firat-2020,mt5}. While it is a convenient solution and often outperforms uniform weighting ($\tau=1$), it reduces the characteristics of languages to their size and reflects the assumption of a zero-sum game in joint training~\citep{mt5}, ignoring more complex interactions~\citep{m2m,wang2021gradient}. Our experiments reveal that even very unbalanced and counter-intuitive schedules can lead to improved results across the board thanks to more intricate and automated sampling. Closest to our work are recent approaches to schedule data based on inter-facet gradient similarity~\citep{wang2020optimizing,wang2020balancing}, which are more computationally expensive.

 \section{Discussion and Conclusion}

We showed that a simple application of the EXP3 algorithm~\cite{auer2002finite, graves2017automated} to the training of a black-box NMT system is a cheaper and non-invasive alternative to task-specific expensively hand-crafted curricula and to heavy RL-based approaches.
Bandit-optimized data usage leads to improved performance compared to the baselines across the board, and sometimes even faster convergence. 
On the difficult task of improving naturalness of translations we gained +0.5--0.9 BLEU on natural on average; on the multi-domain task up to 1.7 points on certain domains using 72\% of the baseline's time to converge; on the multilingual MT task on average---by +1.2 points for translations of 94 languages into English, and by +0.6 points for the reverse.
 
We found intuitive explanations for the learned policies on some of the tasks, but our ability to interpret bandit actions with human reasoning is very limited especially when the number of facets and training steps grow, and also defeats the purpose of replacing possibly flawed human intuitions with learned curricula. As opposed to the expensive development cycles (``train-interpret-retrain'') of post-training data interpretability methods~\cite{koh2020understanding} 
the bandits directly act on their understanding of what is beneficial for the task at hand. After training we can report for each model how much each facet actually mattered, which would increase the transparency of model reporting~\citep{modelcards}, especially for large-scale models~\citep{t5,mt5}. 
 
Finally, there are a few limitations of our approach: Being stateless, unlike RL approaches~\cite{kumar2019reinforcement}, bandits might be short-sighted and keen on exploiting easy data first. Our experiments, though, show that this, with a sufficiently large exploration rate, does not seem to be the case for the tested applications and is not an obstacle to practical use. Another limitation are additional hyperparameters to be set (learning and exploration rates, and reward definitions). Again, we found it relatively easy to navigate in practice by 
stopping unpromising runs early ($\sim$50k steps in our runs, cf.~Figure~\ref{fig:multi_domain_curves}); moreover, the hyperparameters tend to generalize across tasks. We believe that the flexibility provided by the reward definitions would allow to inject domain knowledge and/or signals from potentially multiple objectives, and prior knowledge of the data imbalance could be reflected in the exploration rate. 

Our implementation of EXP3 samples facets in homogeneous batches, but the current SOTA models use heterogeneous ones~\cite{arivazhagan2019massively}. This introduces a limitation and a potential hindrance for optimization~\citep{demystify}, that we hope to address in future work by learning a sampling distribution over individual sentences. With steadily growing training data from more and more sources~\citep{t5,mt5}, it would also be desirable to model facet hierarchies or intersectionalities, e.g., differentiating between domains and translationese vs.~natural within each language pair for a multilingual model.

\section*{Acknowledgements}
We would like to thank Markus Freitag for the help with the setup of the natural vs.~translationese experiments, and Anselm Levskaya and the Flax team for their help with JAX/Flax debugging.


\bibliography{anthology,custom}
\bibliographystyle{acl_natbib}

\appendix

\clearpage
\section{Hyperparameters}\label{sec:hyperparams}
\subsection{Transformer implementation}\label{sec:hapax} 

We abstained from adding the plethora of architecture and pre-processing tweaks common for systems competing in MT benchmarks, and experimented with bare bone Transformer models in order to reduce confounding effects, and keep the code and resulting experiments minimal and clean~\citep{kreutzer2019joey}.\footnote{For our pre-processing pipeline, that is built on top of Tensorflow Datasets, we found that increasing shuffle buffer had a significant positive effect on baseline performance, therefore all experiments were performed for the shuffle buffer size value that was optimal for baselines
.} 

To verify the implementation, we tested it on the WMT14 \texttt{\texttt{en-de}} benchmark, where it scores 27.8 BLEU (without ensembling) vs.~27.3 reported in~\cite{vaswani2017all}.  

\subsection{Natural vs.~translationese} 

We used the same configuration as the \texttt{big} Transformer model from~\cite{vaswani2017all}, except for the MLP dimension which was increased to 8192. Training on TPUv2, we used a learning rate of 0.01, warmup 10000, label smoothing of 0.1, dropout of 0.1 and 96 as batch size.
The maximum length during training was set to 100. Decoding was done with beam size 4 and maximum length 140 during beam search. 

Details on the LM classifier: The first LM was trained on monolingual news crawl data provided by the organizers of the WMT campaign, which comes from news sites originally written in the desired language. The second dataset was generated by (forward) translating data in the source language into the target language by a previously trained MT system. Note that, although we train only on MT generated data, we will use this last LM for identify both human and machine translated data in the training corpus. Our experiments show that this method can help identifying both types of translationese texts, probably due to the fact that MT output exacerbates the characteristics of translationese text. Inspired by \cite{riley-etal-2020}, for each sentence we compare the model score of each of the LMs, and select the class corresponding to the one which produces a better score.

\subsection{Multi-domain}  

Following~\cite{mueller2020domain}, we applied the standard Moses preprocessing pipeline (removing non-printing chars, normalizing punctuation, tokenizing, truecasing and length filtering) to all splits of the data, including the test set. The Subtitles part was limited to 500k sentences and concatenated data was preprocessed jointly with 32,000 BPE merges~\citep{sennrich2016neural}, resulting in a 32,298 vocabulary entries. Maximum training length was 100 post-BPE tokens.

We used the same configuration as the \texttt{base} Transformer model from~\cite{vaswani2017all}. Training on TPUv2 used learning rate 0.01, warmup 4,000, label smoothing 0.1, dropout 0.2 and nominal batch size 256.
Decoding was done with beam size 5 and maximum length 256 during beam search.  
BLEU score were calculated with SacreBLEU on deBPE'ed and detokenized sentences w.r.t.~similarly preprocessed references.

The bandits used the learning rate of 0.1 and exploration rate of 0.25, found by grid search over the range [0.001, 0.01, 0.1] and [0.5, 0.25, 0.1] respectively.

\subsection{Multilingual} 

\paragraph{TED} For TED we train the models on 4 V100 GPUs with a batch size of 64 sentences for 50k steps, a warmup period of 4k steps for a learning rate schedule with linear increase and square-root decay and a base learning rate of 0.0625. Training sentences up to a length of 512 tokens are considered. For inference, beam width is set to 4. Models are validated every 2k steps.
for OPUS100 5. Bandit learning and exploration rate were tuned over a grid search over the range [0.001, 0.01, 0.01] and [0.1, 0.2, 0.3, 0.4, 0.5] respectively, with training up to 10k training steps. For the \texttt{diverse} task the best setting was (0.1, 0.3) and for the \texttt{related} task (0.01, 0.2).

\paragraph{OPUS100} For OPUS the models are trained with a total batch size of 256 sentences, 1k warmup steps and the same learning rate schedule as for TED. For inference we use beams of width 5. Models are trained for 500k steps and validated every 8k.
The best configuration of bandit learning and exploration rate is (0.01, 0.5) for all settings (M2O, O2M, M2M).

\paragraph{SentencePiece} For balanced subword representations we upsample all languages to the maximum size across languages and then using the SP option \texttt{large\_corpus} to subsample uniformly from their concatenation.

\begin{table*}[h!]
    \centering
    \begin{tabular}{l|cccc|cccc}
    \toprule
         & \multicolumn{4}{c|}{\textbf{M2O}} & \multicolumn{4}{c}{\textbf{O2M}} \\
        & All & Low & Med & High & All & Low & Med & High \\
    \midrule
       Base  & 21.31 & 25.94 & 20.10 & 19.90 & 18.36 & 15.64 & 17.78 & 20.00  \\
       \texttt{dev-pgnorm} & 21.93 & 26.09 & 20.61 & 20.81 & 19.58 & 19.36 & 17.96 & 20.69 \\
    \bottomrule
    \end{tabular}
    \caption{\textbf{M2M:} Avg BLEU across languages for OPUS100's 94 test sets grouped by training corpus size as in~\citep{zhang2021share}.}
    \label{tab:opus100_m2m}
\end{table*}
\section{OPUS100 M2M}\label{app:m2m}
Table~\ref{tab:opus100_m2m} lists the result for many-to-many translation for the OPUS100 benchmark.

\section{Multi-lingual bandit strategies}\label{app:strategies}

\begin{figure}
    \centering
    \includegraphics[width=\columnwidth]{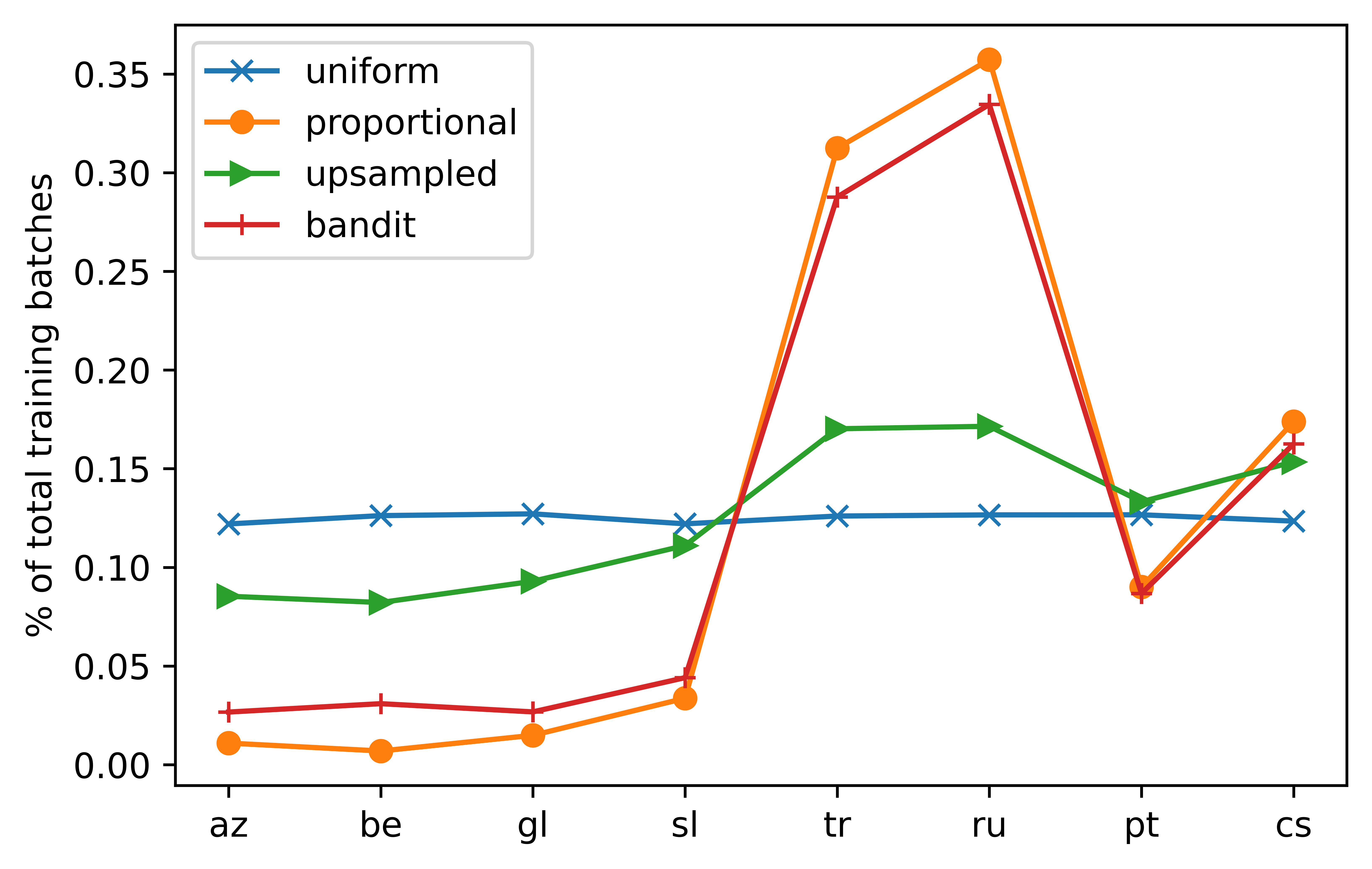}
    \caption{Ratio of training batches from each language throughout training for TED-related M20.}
    \label{fig:ted_related_plays}
\end{figure}

Figure~\ref{fig:ted_related_plays} shows the ratio of training batches from each language, averaged across the complete training run. The corresponding diagram for the diverse subset is in Figure~\ref{fig:probs_m2o_diverse}. 
Again, we find that the bandit mimics the proportional sampling strategy, with slight upsampling of the lowest-resource pairs.

Table~\ref{tab:train_batches} lists the top 5 sampled languages for each OPUS setting. For M2O these are largely high-resource pairs, for O2M low-resource pairs, and for M2M pairs with English as target were generally sampled more, but the top 5 are a mix of high- and resource languages.

\begin{minipage}[b][10em]{\columnwidth}
~
\end{minipage}

\begin{table}[t!]
    \centering
    \resizebox{\columnwidth}{!}{
    \begin{tabular}{lcccc}
    \toprule
   Setting &  Lang. & \% Train. Batches & Train. Size & Test BLEU\\
    \midrule
    \multirow{5}{*}{M2O} &  \texttt{nl-en}    &  10.7 & 1M & 28.86\\
        & \texttt{cs-en} & 3.5 & 1M & 27.39\\
        & \texttt{ms-en} & 3.0 & 1M & 27.32\\
        & \texttt{sh-en} & 2.9 & 267k & 28.30\\
        & \texttt{sr-en} & 2.8 & 1M & 57.21\\
    \midrule 
     \multirow{5}{*}{O2M}  & \texttt{en-fy}    &  5.9 & 54k & 35.19 \\
        & \texttt{en-ga} & 4.8 & 290k & 12.20 \\
        & \texttt{en-ky} & 4.7 & 27k & 18.72 \\
        & \texttt{en-mg} & 3.6 & 591k & 16.52 \\
        & \texttt{en-ug} & 3.6 & 72k & 9.61 \\
    \midrule
    \multirow{5}{*}{M2M} & \texttt{as-en}    &  4.3 &  138k & 29.24 \\
        & \texttt{ta-en} & 2.5 & 227k & 0.91 \\
        & \texttt{li-en} & 2.1 & 26k & 51.28 \\
        & \texttt{fa-en} & 2.0 & 1M & 20.28\\
        & \texttt{da-en} & 1.4 & 1M & 19.81 \\
    \bottomrule
    \end{tabular}
    }
    \caption{Top 5 sampled languages for M2O, O2M, and M2M OPUS.}
    \label{tab:train_batches}
\end{table}

\begin{minipage}[b][20em]{\columnwidth}
~
\end{minipage}

\clearpage
\onecolumn
\section{Bandit arm probabilities}\label{app:probs}

\begin{figure*}[h]
    \centering
    \resizebox{\textwidth}{!}{\input{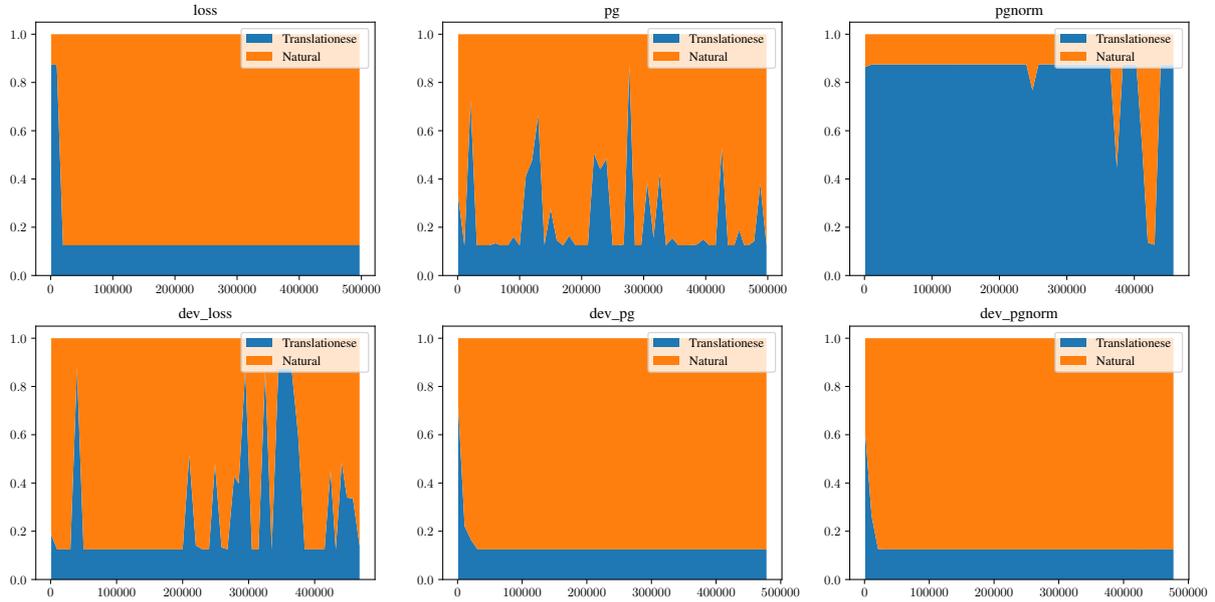}}
    
    \caption{Evolution of probabilities during training on the WMT \texttt{en-de} task (without CDS filtering). See \S\ref{sec:nat2trans} for interpretation.
    }
    \label{fig:wmtbase_ende_base_probs}
\end{figure*}

\begin{figure*}[h]
    \centering
    \resizebox{\textwidth}{!}{\input{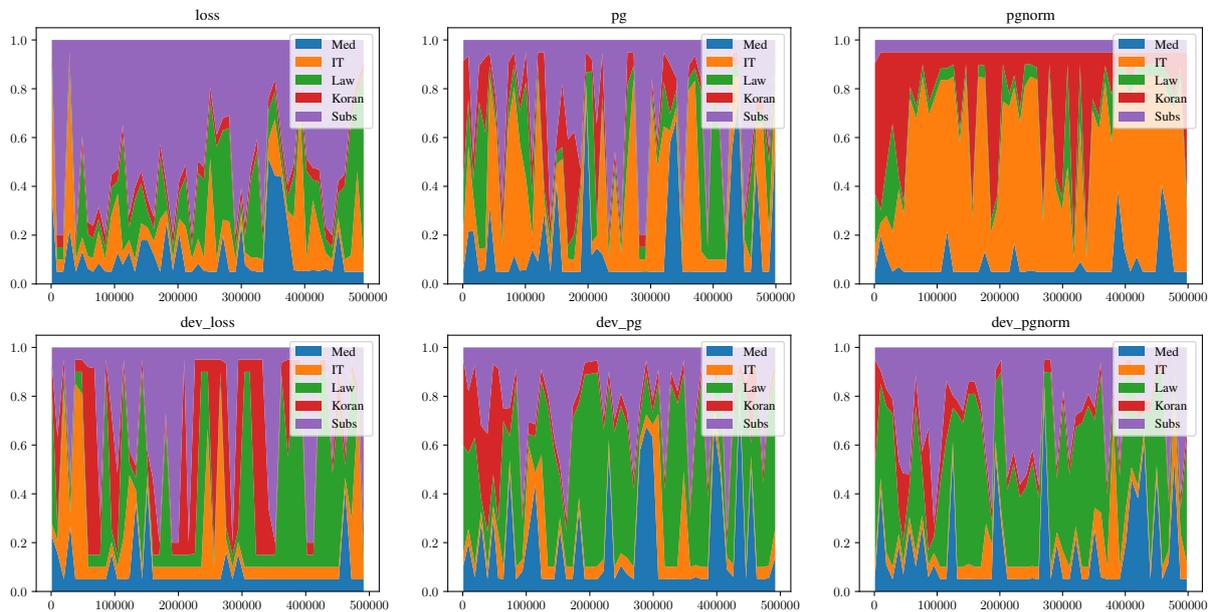}}
    
    \caption{Evolution of probabilities during training on the multi-domain task. See \S\ref{sec:domain} for interpretation.}
    \label{fig:multi_domain_probs}
\end{figure*}

In Figures~\ref{fig:wmtbase_ende_base_probs} and~\ref{fig:multi_domain_probs} we plot the evolution of bandit arm (facet) sampling probabilities over time to illustrate the learned curricula for the \texttt{en-de} natural vs.~translationese and multi-domain tasks. Best viewed in color.

\clearpage

\end{document}